\pdfoutput=1

\documentclass[11pt]{article}
\usepackage{graphicx}  
\usepackage[preprint]{acl}
\usepackage{array}
\usepackage{xcolor}
\definecolor{softpink}{RGB}{233,185,193}   
\definecolor{softblue}{RGB}{160,180,200}   

\definecolor{entitypink}{RGB}{255,192,203} 
\usepackage{subcaption}
\usepackage{tikz}
\usepackage{enumitem} 
\usetikzlibrary{arrows.meta, positioning, decorations.pathreplacing}
\usepackage{makecell}
\usepackage{multirow}
\usepackage{graphicx}
\usepackage{booktabs}
\usepackage{amssymb} \usepackage{float}
\usepackage{xcolor}
\usepackage[table]{xcolor} 
\usepackage{adjustbox}  
\usepackage{times}
\usepackage{latexsym}
\usepackage{linguex}

\usepackage[T1]{fontenc}

\usepackage[utf8]{inputenc}

\usepackage{microtype}

\usepackage{inconsolata}

\usepackage{graphicx}

\title{DiscoTrack: A Multilingual LLM Benchmark for Discourse Tracking}

\author{
Lanni Bu \quad Lauren Levine \quad Amir Zeldes \\
Georgetown University \\
\texttt{\{lb1437, lel76, amir.zeldes\}@georgetown.edu}
}

\begin{document}
\maketitle
\begin{abstract}
Recent LLM benchmarks have tested models on a range of phenomena, but are still focused primarily on natural language understanding for extraction of explicit information, such as QA or summarization, with responses often targeting information from individual sentences. We are still lacking more challenging, and importantly also multilingual, benchmarks focusing on implicit information and pragmatic inferences across larger documents in the context of discourse tracking: integrating and aggregating information across sentences, paragraphs and multiple speaker utterances. To this end, we present DiscoTrack, an LLM benchmark targeting a range of tasks across 12 languages and four levels of discourse understanding: salience recognition, entity tracking, discourse relations and bridging inference. Our evaluation shows that these tasks remain challenging, even for state-of-the-art models.
\end{abstract}

\section{Introduction}
A growing number of benchmarks, such as GLUE \citep{wang2018glue} and CLUE \citep{xu2020cluecorpus2020}, have been developed to evaluate LLMs across a wide range of phenomena, languages, and task types. However, the majority of existing benchmarks focus on extracting explicit information, often from individual sentences, and offer limited coverage of discourse-level phenomena that require reasoning across multiple sentences or larger contexts. Some recent efforts, such as Disco-Bench \citep{wang2023discobench}, have begun to address discourse-level evaluation. However, this benchmark is limited to only two languages and still focuses on QA and translation, rather than tracking discourse in terms of entities mentioned or implicit relations between utterances.

This gap in coverage motivates our work, DiscoTrack. In this paper, we present a multilingual benchmark designed to evaluate large language models’ capabilities in discourse tracking: the ability to gather information that is not explicitly stated from across long documents to identify relations and group information based on discourse structure. 

The main contributions of this paper are:

\begin{itemize}
\setlength\itemsep{0.1em} 
    \item Releasing an openly available, multilingual benchmark, covering 12 typologically diverse languages: English, Chinese, Czech, French, German, Italian, Polish, Portuguese, Russian, Spanish, Thai and Turkish.
    \item Broadening the range of phenomena for LLM benchmarking to include salience, entity tracking, discourse relations, and multiple aspects of bridging anaphora.
    \item Presenting a thorough evaluation of several LMs and corresponding human performance specifically on tasks at the discourse level.
    \item Providing error analysis to expose specific model strengths and weaknesses, with confusion matrices, comparing zero and few-shot settings, detailed separate scores on English and lower-resource lanugages, as well as comparisons to human errors.
\end{itemize}

\section{Related work}

The most closely related effort to our work is Disco-Bench \cite{wang2023discobench}, which specifically targets some discourse phenomena. However Disco-Bench is limited to Chinese and English, and targets discourse rather indirectly: for discourse understanding, it only uses three monolingual Chinese tasks: speaker identification, zero anaphora reconstruction (which is much less relevant for languages such as English) and classic reading comprehension questions. For generation it uses a mix of machine translation, text expansion and language modeling (text filling/completion), all of which benefit from discourse understanding, but do not probe discourse processing capabilities directly.

While both Disco-Bench and our work focus on evaluating discourse-aware capabilities of LLMs, the broader field of LLM evaluation encompasses a wide range of efforts that test different linguistic phenomena using diverse models, tasks, and prompting strategies. Recent benchmarking efforts reveal two notable trends: a shift from monolingual to multilingual evaluation, and a progression from simple sentence-level tasks toward more complex, cognitively demanding language understanding.

This progression in task complexity is reflected in a variety of recent benchmarks. Early efforts such as GLUE and SuperGLUE \citep{wang2019superglue,wang2023discobench} focused on localized classification tasks like natural language inference (NLI) and coreference linking. Later benchmarks like MMLU \citep{hendrycks2020measuring}, LM Evaluation Harness \cite{eval-harness} and BIG-Bench \citep{kazemi2025big} expanded the scope to include general-purpose reasoning, knowledge retrieval, and problem-solving across domains. MT-bench \citep{bai2024mt} and LiveBench \citep{white2024livebench} have further shifted attention toward open-ended generation, assessing coherence and fluency in dialogue or multi-turn responses. Some efforts, such as AGIEval \citep{zhong2023agieval}, explicitly model human-like cognitive testing by evaluating performance on standardized exam questions. While many of these benchmarks and tasks involve reasoning over multiple sentences, they are not well suited to isolating discourse capabilities in the sense of computational discourse processing \cite{Stede2012}, i.e.~with a focus on revealing implicit structure, priorities and unexpressed relationships that form the backbone of coherent texts.

\section{DiscoTrack}
\begin{table*}[t]
\centering

\scriptsize
\setlength{\tabcolsep}{3pt}
\renewcommand{\arraystretch}{1.2}

\begin{tabular}{lcccccccccccc}
\toprule
\textbf{Task Group} & \textbf{English} & \textbf{Czech} & \textbf{French} & \textbf{German} & \textbf{Portuguese} & \textbf{Russian} & \textbf{Spanish} & \textbf{Italian} & \textbf{Turkish} & \textbf{Chinese} & \textbf{Thai} & \textbf{Polish} \\
\midrule
BRIDGDET   & 466 &  &  &  &  &  &  &  &  &  &  &  \\
BRIDGTYPE   & 219 &  &  &  &  &  &  &  &  &  &  &  \\
BRIDGANTE    & 195 &  &  &  &  &  &  &  &  &  &  &  \\
\midrule
SALTOP    & 207 &  &  &  &  &  &  &  &  &  &  &  \\
SALREL    & 500 &  &  &  &  &  &  &  &  &  &  &  \\
\midrule
RELCLF & 500 & 500 & 500 & 500 & 500 & 500 & 500 & 500 & 500 & 500 & 500 & 500 \\
RELEXP & 500 & 500 &  & 322 & 500 &  &  & 262 & 500 & 500 & 500 & 500 \\
RELIMP & 500 &  &  & 301 & 316 & 296  &  &  & 342 & 500 &  & 326 \\
\midrule
ENTMAX   & 234 & 500 & 143 & 221 &  & 179 & 500 &  & 22 &  &  & 500 \\
ENTCOUNT & 405 & 500 & 413 & 173 &  & 373 & 472 &  & 376 &  &  & 500\\
ENTCOREF & 500 & 500 & 500 & 104 &  & 500 & 500 &  & 500 &  &  & 500\\
\bottomrule
\end{tabular}
\label{tab:tasks-availability}

\caption{Language Availability and Number of Instances per Task in the DiscoTrack Benchmark.}
\label{tab:tasks-availability}
\end{table*}

\begin{figure}
    \centering
\begin{tikzpicture}[>=Stealth, node distance=1cm, font=\small]
\tikzset{word/.style={inner sep=0.5pt, outer sep=0pt, anchor=base}}
\node[word] (k) at (0,0) {\textbf{Kim}};
\node[word, right=0mm of k, xshift=1.5mm] (and) {and};
\node[word, right=0mm of and, xshift=1.5mm] (m1) {\textcolor{softpink}{\textbf{Mary}}};
\node[word, right=0mm of m1, xshift=1.5mm] (w) {went};
\node[word, right=0mm of w,  xshift=1.5mm] (t) {to};
\node[word, right=0mm of t,  xshift=1.5mm] (s) {school};
\node[word, right=0mm of s,  xshift=1.5mm] (wi) {with};
\node[word, right=0mm of wi, xshift=1.5mm] (j) {\textcolor{lightgray}{Jamie}.};

\draw[teal, thick, rounded corners=2pt] 
  ([xshift=-1pt, yshift=-2pt]k.south west) rectangle ([xshift=1pt, yshift=2pt]m1.north east);

\draw[teal, thick, rounded corners=2pt] 
  ([xshift=-1pt, yshift=-2pt]s.south west) rectangle ([xshift=1pt, yshift=2pt]s.north east);

\draw[->, thick, teal] (s.north) .. controls +(0,1) and +(0,1) .. node[above, midway, font=\footnotesize] {bridging: entity-associative} (m1.north);
\node[below=0.5cm of k.base west, anchor=base west] (sud) {[Suddenly, \textcolor{softpink}{\textbf{Mary}} fell.]};
\node[right=0.05cm of sud, anchor=base west] (kim) {[\textbf{Kim} pushed \textcolor{softpink}{\textbf{her}}!]};
\draw[{Stealth[length=3mm]}-, thick, cyan]
  (sud.south) .. controls +(0,-1) and +(0,-1)
  .. node[below, midway, font=\footnotesize] {cause + precedence}
  (kim.south);
\end{tikzpicture}
    \caption{Sample discourse illustrating associative bridging anaphora, implicit \textit{cause} and \textit{precedence} discourse relations, tracking the entity Mary, and relative salience, where the less salient `Jamie' is shown in gray.}
    \label{fig:discourse-example}
\end{figure}
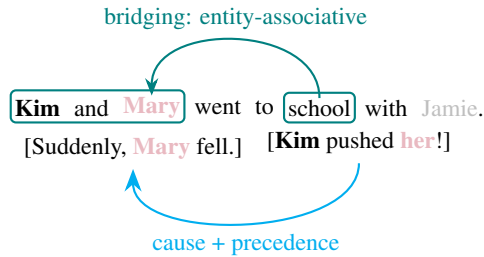

In order to specifically evaluate LLM discourse capabilities, DiscoTrack focuses on tasks which target multi-sentential context, long distance dependencies, and whole document understanding. 
Our benchmark covers a wide range of languages (see Table~\ref{tab:tasks-availability}), enabling cross-lingual evaluation.
We select tasks in four areas:

\begin{itemize}
\setlength\itemsep{0.1em} 
    \item \textbf{Salience recognition} - by focusing on the distinction between more and less important entities in a document, we force models to relativize or rank some or all entities mentioned in the discourse simultaneously. For example in Figure \ref{fig:discourse-example}, Kim and Mary (in bold) are more salient than Jamie (in gray).
    \item \textbf{Entity tracking} - as discourse develops, entities are referred back to, and models must distinguish between newly introduced concepts and ones that tie back to the preceding discourse. For example, models should notice that there is only one Mary, and she is mentioned 3 times in Figure \ref{fig:discourse-example}.
    \item \textbf{Discourse relations} - the relationship between multiple propositions is often left unexpressed, and here we force models to identify, interpret and complete information signaling such relations. In Figure \ref{fig:discourse-example}, these include temporal precedence (\textit{first} Kim pushed Mary, \textit{then} she fell) and causality (Mary fell \textit{because} of Kim pushing).
    \item \textbf{Bridging inference} - we challenge models to explicitly identify implicit relationships between mentioned concepts, including unexpressed part-whole, set-member and other associative links which make the nature of newly introduced entities inferable in context. For example in Figure \ref{fig:discourse-example} we infer that the school is the one Kim and Mary regularly go to.
\end{itemize}

Each area corresponds to a module in the benchmark and includes 2--3 sub-tasks, described below.

\subsection{Salience recognition}

Salience in language can refer to a range of concepts (see \citealt{BoswijkColer2020}), but for the present purpose, we define salience as the relative degree of prominence or importance that can be assigned to concepts in a document, which can be identified by a number of methods. In particular, our focus in this module of the benchmark is on Salient Entity Extraction (SEE, \citealt{lin-zeldes-2024-gumsley}).

SEE datasets have taken a range of approaches for identifying the most relevant entities in context, including human annotation \cite{dojchinovski2016crowdsourced}, extraction from click-stream data \cite{gamon2013identifying}, using hyperlinks and categories in Wikipedia data \cite{wu2020wn} and automatically aligning entities extracted from documents using Named Entity Recognition (NER) with the same entities mentioned in summaries of those documents \cite{dunietz2014new}. Unfortunately, almost all work on SEE has been conducted on English, meaning that in our benchmark this area is currently restricted to the evaluation of English language prompts only.

For our salience module, we employ a recently developed dataset annotated with graded salience scores, which range from 0--5 \cite{lin2025gumsage}. The scores were computed by obtaining five summaries for each document in the freely available GUM corpus \cite{zeldes2017gum}, which contains named and non-named entity and coreference annotations, and then manually aligning entities mentioned in the summaries to the document annotations. Entities mentioned in all five summaries receive the maximum score of 5, while entities mentioned in no summary score 0, using the reasoning that summaries will tend to mention the most salient entities in each document. The tasks in this module are: 

\paragraph{Top salient (\textsc{saltop})} Given an entire document, extract the top 3 most salient entities (all documents in the dataset have at least 3 salient entities). The model must return a string representation corresponding to exactly three mentions of entities scoring the maximum score in the document. Since we have gold coreference annotations, we accept any non-pronominal mention of any subset of 3 entities corresponding to the top scores. For example, if there are four entities with the score of 5, we accept any response with a single mention from three of those mention clusters. If there are 2 mentions with score 5 and 3 with score 4, we accept any three mentions from 3 of the 5 distinct clusters, etc.

\paragraph{Relative salience (\textsc{salrel})} Given a document and two entity mentions, models must provide a binary response indicating whether the first entity is more salient than the second. We use the gold standard salience scores to select pairs of entity mentions. Specifically, we first filter out all mentions with a salience score of 0 in order to avoid overly trivial contrasts between highly salient and entirely non-salient entities.
Instead, we focus on more challenging comparisons by constructing balanced pairs from entity mentions with scores ranging from 1 to 5.

\subsection{Entity tracking}
\begin{figure*}[t]
  \vspace{-60pt} 
  \centering
  \includegraphics[width=\textwidth]{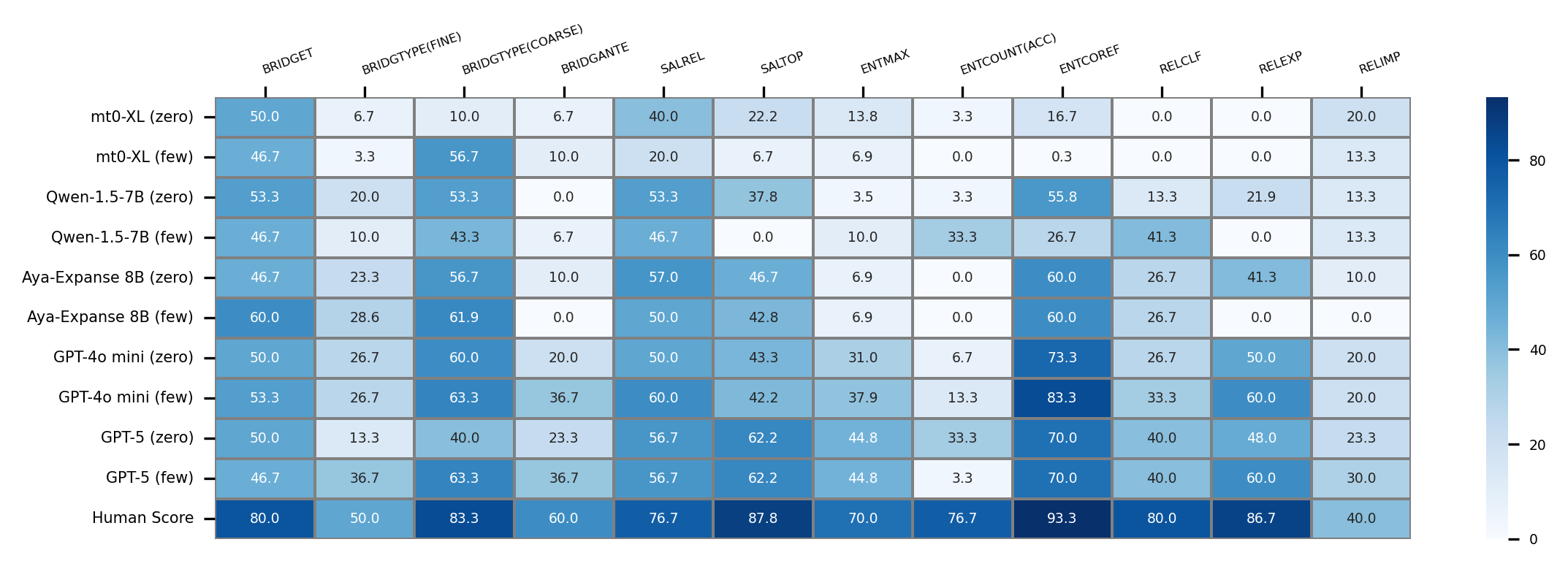}
  \vspace{-20pt} 
  \caption{LLM and human performance (accuracy \%).}
  \label{fig:llm-human-performance}
\end{figure*}

Entity tracking corresponds broadly to the well-studied task of coreference resolution \cite{pradhan-etal-2014-scoring}, in which spans of text that refer to the same entity in the world (e.g.~`Mary' and `her' in Figure \ref{fig:discourse-example}) are grouped into clusters. However, while structured coreference resolution can rely on pairwise linking each mention to just one antecedent to form clusters, LLMs reading a document holistically must implicitly track which expressions refer to the same entities in order to make correct inferences about discourse content. Rather than use structured coreference, which is a fairly unnatural task for text-driven LLMs, we employ a number of reasoning tasks more similar to a question-answering paradigm, which lend themselves to modeling with LLMs while forcing models to attend to entire documents.

Coreference information is available in a wide range of languages, and has recently been made available in the multilingual CorefUD dataset \cite{nedoluzhko-etal-2022-corefud} in a uniform format, and we rely on this dataset to extract the subtask data for this module. The tasks are:

\paragraph{Top mentioned entity recognition (\textsc{entmax})} In this task the objective is to extract the entity mentioned the most times in the input document. Note that this combines aspects of mention identification, coreference resolution, and aggregate reasoning over the discourse. To design the task, we use the coreference clusters in the gold standard CorefUD data to count the mentions in each cluster and identify the most mentioned cluster. If the cluster consists entirely of pronouns (for example repetitions of `I' and `me' in a long monologue), we accept any form of mention from the cluster as a correct answer. If the cluster contains any non-pronominal mentions, we accept any member of the set of those non-pronominal mentions which is unambiguous across clusters and forms a complete phrase. For example, if the most mentioned entity in the document is former US President Barack Obama, who is mentioned repeatedly as `U.S. President Barack Obama', `President Obama' or `he', we accept only `U.S. President Barack Obama' and `President Obama', but not `Obama' (as such a mention did not appear in the discourse) and not `he' (as it is a pronoun). If multiple presidents are mentioned as `the President', we also do not accept that response. Prompts in the experiments instruct models to follow these guidelines, and our human evaluation follows the same rules.

\paragraph{Entity mention count extraction (\textsc{entcount})} This task is similar to the last, but instead of picking the most common entity given the document, we now name a specific entity using a single, non-ambiguous, non-pronominal mention (i.e.~a mention whose string form is found in only one cluster and is not a pronoun), and we ask exactly how often it is mentioned, including as a pronoun. In order to balance frequent and rare entities, and to make the task tractable to humans, we evenly balance entities appearing between 1 and 20 times in the documents. For example, if the document consisted of the example in \ref{ex:ent-count} and the prompt asked for a count of mentions for the entity mentioned as `Kim', the answer would be 4.

\ex. 
\textit{\textbf{Kim} said \textbf{she} liked to take photos of \textbf{herself} with \textbf{her} cat.}\label{ex:ent-count}\\
\textbf{Mentions of ‘Kim’}: \textit{4}

\begin{table*}[t]
\centering
\setlength{\tabcolsep}{2pt}
\renewcommand{\arraystretch}{1.05}
\scriptsize
\begin{adjustbox}{max width=\textwidth}
\begin{tabular}{
  l|
  c cc c|c c|
  ccc ccc ccc|
  ccc ccc
  ccc
}
\toprule
\multirow{2}{*}{\textbf{Model / Setting}} &
\multirow{2}{*}{\textbf{BRIDGDET}} &
\multicolumn{2}{c}{\textbf{BRIDGTYPE}} &
\multirow{2}{*}{\textbf{BRIDGANTE}} &
\multirow{2}{*}{\textbf{SALREL}} &
\multirow{2}{*}{\textbf{SALTOP}} &
\multicolumn{3}{c}{\textbf{ENTMAX}} &
\multicolumn{3}{c}{\textbf{ENTCOUNT}} &
\multicolumn{3}{c}{\textbf{ENTCOREF}} &
\multicolumn{3}{c}{\textbf{RELCLF}} &
\multicolumn{3}{c}{\textbf{RELEXP}} &
\multicolumn{3}{c}{\textbf{RELIMP}} \\
\cmidrule(lr){3-4}
\cmidrule(lr){8-10} \cmidrule(lr){11-13} \cmidrule(lr){14-16}
\cmidrule(lr){17-19} \cmidrule(lr){20-22} \cmidrule(lr){23-25}
& & \textbf{fine} & \textbf{coarse} & & & &
\textbf{All} & \textbf{Eng} & \textbf{Non} &
\textbf{All} & \textbf{Eng} & \textbf{Non} &
\textbf{All} & \textbf{Eng} & \textbf{Non} &
\textbf{All} & \textbf{Eng} & \textbf{Non} &
\textbf{All} & \textbf{Eng} & \textbf{Non} &
\textbf{All} & \textbf{Eng} & \textbf{Non} \\
\midrule

mt0-XL (zero)
  & 45.49 & 6.85 & 12.33 & 2.05 & 54.2 & 23.3
  & 22.47 & 28.21 & 21.66
  & 2.85 & 7.4 & 2.204
  & 38.18 & 38.4 & 38.15
  & 4.52 & 2.6 & 4.69
  & 0 & 0 & 0
  & 19.90 & 18.6 & 20.12
  \\

mt0-XL (few)
  & 50.21 & 1.37 & 47.03 & 5.13 & 13.8 & 4.8
  & 2.69 & 2.99 & 2.65
  & 4.11 & 7.4 & 4.59
  & 42.87 & 49 & 41.99
  & 3.38 & 0.4 & 3.65
  & 0.13 & 0 & 0.13
  & 17.29 & 12.6 & 18.08
  \\

Qwen-1.5-7B (zero)
  & 48.28 & 30.14 & 47.95 & 0.51 & 52.8 & 35.1
  & 13.49 & 22.22 & 12.24
  & 6.65 & 5.19 & 6.85
  & 54.43 & 55.8 & 57.66
  & 6.48 & 10.6 & 6.11
  & 7.09 & 16.93 & 5.69
  & 14.17 & 14.8 & 14.06
  \\

Qwen-1.5-7B (few)
  & 52.15 & 15.98 & 49.4 & 2.56 & 59.2 & 41.22
  & 11.83 & 23.5 & 10.17
  & 5.73 & 8.4 & 5.35
  & 45.06 & 41.4 & 45.58
  & 5.28 & 10.4 & 4.82
  & 18.07 & 18.32 & 18.03
  & 6.83 & 15 & 5.47
  \\

Aya-Expanse 8B (zero)
  & 50.21 & 29.68 & 47.03 & 8.21 & 52.2 & 38.65
  & 10.08 & 25.21 & 7.91
  & 2.77 & 3.95 & 2.60
  & 55.37 & 54.2 & 55.54
  & 3.38 & 5.4 & 3.2
  & 0 & 0 & 0
  & 13.16 & 8.2 & 13.98
  \\

Aya-Expanse 8B (few)
  & \textbf{62.23} & 29.69 & 47.03 & 1.03 & 49.6 & 39.45
  & 12.09 & 29.91 & 9.54
  & 3.59 & 5.19 & 3.36
  & 51.49 & 50 & 51.71
  & 14.72 & 21.2 & 14.13
  & 9.09 & 0 & 9.09
  & 4.01 & 0.4 & 4.21
  \\

GPT-4o mini (zero)
  & 57.30 & \textbf{32.42} & 54.34 & 18.97 & 52.6 & 39.61
  & 38.73 & 31.62 & 39.74
  & \textbf{8.76} & 10.37 & \textbf{8.53}
  & 69.70 & 76 & 68.80
  & 28 & 33.4 & 27.51
  & 48.60 & 47.18 & 48.80
  & 13.97 & 21 & 12.80
  \\

GPT-4o mini (few)
  & 49.14 & 30.59 & 52.97 & \textbf{42.05} & 67.0 & 43.2
  & \textbf{41.48} & 35.04 & \textbf{42.40}
  & 8.06 & \textbf{11.60} & 7.55
  & 72.97 & \textbf{81.6} & 71.73
  & 28.57 & 41.4 & 27.4
  & \textbf{50.37} & \textbf{52.7} & \textbf{50}
  & 19.47 & \textbf{24.4} & 19.48
  \\

GPT-5 (zero)
  & 53.65 & 21.92 & 46.56 & 38.46 & 54.2 & 56.2
  & 34.45 & \textbf{37.18} & 34.06
  & 8.1 & 10.37 & 7.78
  & 60.71 & 72 & 59.1
  & \textbf{48.02} & \textbf{45.8} & \textbf{48.02}
  & 39.59 & 46.98 & 38.66
  & 11.74 & 21.2 & 10.16
  \\

GPT-5 (few)
  & 54.29 & 31.51 & \textbf{58.9} & 38.97 & \textbf{67.8} & \textbf{59.26}
  & 35.97 & 32.19 & 36.51
  & 8.08 & 11.36 & 7.62
  & \textbf{73.22} & 75.4 & \textbf{72.91}
  & 38.37 & 40.2 & 38.2
  & 42.94 & 47.18 & 42.42
  & \textbf{23.33} & \textbf{24.4} & \textbf{23.15}
  \\

\bottomrule
\end{tabular}
\end{adjustbox}
\caption{Micro scores (per task).
\label{tab:pertask}
\textbf{Bold} numbers indicate the best (highest) value per task.}
\end{table*}
\paragraph{Adversarial coreference resolution (\textsc{entcoref})} In this task, two mentions are highlighted within curly brackets -- first the antecedent (`mass strikes' in example \ref{ex:ent-bin}) and then the anaphor (`it').

\ex.
\textit{We’ve seen \{mass strikes all around the world\}, in countries that we wouldn’t expect \{it\}.} \\
\textbf{Are these mentions coreferent? } \textit{Yes}\label{ex:ent-bin}

The objective is to decide whether the two entities refer to the same thing in the world or not. In order to make the task more challenging than in previous benchmarks incorporating coreference, we pair true positives with `confusing' negative examples, which we extract by running a recent multilingual neural coreference resolution system, MSCAW-coref \cite{liu-etal-2024-mscaw}, on our documents, which is available as part of the Stanza NLP package \cite{qi-etal-2020-stanza}. We obtain negative pairs by selecting an anaphor predicted by the system for which we know a gold standard mention exists, and then select the highest probability \textit{incorrect} antecedent considered by the system, again verifying that its mention exists in gold data. The filter for gold mentions ensures that no very easy cases are included which can be ruled out based on incorrect mention boundaries. A negative example is shown in \ref{ex:ent-bin-neg}.

\ex. 
\textit{\{These themes\} fall under two broad categories: \{negotiation (i.e., emotional responses) and tactical negotiation (i.e., cognitive, or reasoned responses)}.\} \\
\textbf{Are these mentions coreferent? }\textit{No}\label{ex:ent-bin-neg}

Although the categories of `These themes' are closely related to the themes themselves, as shown by the system's confusion in resolving the reference, they are in fact distinct entities. Full document context is provided for all examples.

\subsection{Discourse relations}

Multilingual identification of implicit and explicitly marked discourse relations such as \textsc{cause} or \textsc{concession}, and identifying connective words associated with such relations (e.g.~\textit{because} can mark \textsc{cause} as well as \textsc{explanation}) are crucial capabilities for the purpose of discourse understanding and explainable NLP \cite{liu-etal-2023-hits,saeed-etal-2025-implicit}. To test models' ability to interpret discourse relations and their markers in text correctly, we assemble multilingual data for three text-based tasks, which we derive from the DISRPT shared task's discourse relation classification and connective detection multilingual datasets \cite{braud-etal-2024-disrpt}, as well as a number of additional datasets annotated according to the Penn Discourse Treebank (PDTB) framework \cite{prasad-etal-2018-discourse}. The tasks are as follows:

\paragraph{Relation classification (\textsc{relclf})} The model is presented with a document and two spans of text from within it, each of which may be discontinuous, and is tasked with selecting the correct relation from an inventory of 17 possible labels (see Appendix~\ref{sec:label-appendix} for the full list of labels). Valid solutions correspond to a single label (the data excludes cases with multiple labels), and we use accuracy as the metric. An English example input (excluding the prompt with whole document context and list of possible labels) for the two argument spans (arg1 and arg2) and output are given in \ref{ex:relclf}.

\ex. Arg1: \textit{With the development of artificial twines the sisal boom ended,} \\
Arg2: \textit{and Mérida slowed to a more sleepy provincial capital} \\
\textbf{Discourse relation:}
\textit{causal} \label{ex:relclf}

Because we can interpret the text to mean that `the sisal boom ended and (\textbf{as a result}) M\'{e}rida slowed', the correct relation type is \textsc{causal}.

\paragraph{Explicit connective detection (\textsc{relexp})} The model is presented with a sentence which may or may not include a connective such as `but', `on the other hand' etc. The inventory of possible connectives for each language is provided based on the DISRPT shared task training data, and the model must return the input sentence with any instances of connectives surrounded by curly brackets if present, as shown in example \ref{ex:explicit}, or the string `no explicit connective' if no connective appears, as in \ref{ex:explicit-negative}.

\ex. \label{ex:explicit}
\a. 
\textit{If you prefer larger cupcakes, then add a half a cup}
\b. Output: \textit{\{If\} you prefer larger cupcakes, \{then\} add a half a cup}

\ex. \label{ex:explicit-negative}
\a. 
\textit{She achieved prominence as the star of Elevator to the Gallows (1958)}
\b. Output: \textit{no explicit connective}

\paragraph{Implicit connective prediction (\textsc{relimp})} In this task, the model is presented with two consecutive sentences (again, arg1 and arg2) that are connected by an implicit discourse relation, but for which no explicit connective appears in the text. The model is tasked with reconstructing the most appropriate connective that a human would use to connect the two sentences, making the relation explicit, as in \ref{ex:implicit}. Negative examples are also provided, for which no connective is appropriate, and to which the correct response is `no implicit connective', as in \ref{ex:implicit-negative} (examples derived from GUM, \citealt{zeldes2017gum}). 

\ex. Arg1: \textit{The French then moved} \\
Arg2: \textit{to capture Mexico City} \\
\textbf{Implicit connective}: \textit{in order} \label{ex:implicit}

\ex. Arg1: \textit{Leah} \\
Arg2: \textit{She snoozing on the floor?} \\
\textbf{Implicit connective}: \textit{no implicit connective} \label{ex:implicit-negative}

\subsection{Bridging inference}

Bridging is an anaphoric phenomenon where the referent of a newly introduced entity is inferable due to the understanding of a previous entity in the discourse. For instance, in the example "I saw a house. The door was red.", "the door" is understood to specifically be the door of the aforementioned "house", due to a semantic part-whole relationship between the bridging anaphor "the door" and its associative antecedent "a house". However, there are many ways that instances of bridging can manifest in a discourse beyond part-whole relations, including more general associative relationships. 

Bridging resolution is a task where bridging anaphora are automatically detected and resolved back to their associative antecedents. Thus far, bridging resolution has proven to be a difficult task, with SoTA systems achieving F-scores of < 0.5 on reference resources \cite{kobayashi-etal-2023-pairspanbert}. Still, for an LLM to have full understanding of a discourse, an understanding of the implicit, associative relationships between bridging anaphora and their antecedents is essential. 

For our benchmark evaluation, we use bridging annotations from the test set of the GUMBridge corpus \cite{levine2025subjectivityannotationbridginganaphora}, which each consist of a bridging anaphor and its associative antecedent. Each bridging pair in the test set is also categorized with a subtype annotation from the bridging subtype classification schema developed for GUMBridge (see the Appendix~\ref{sec:label-appendix} for the subtypes). 

\paragraph{Bridging detection (\textsc{bridgdet})}
The model is presented with a document containing several noun phrases and must decide
whether a given noun phrase is a \emph{bridging anaphor} — i.e., a noun phrase that refers to something
associated with, but not identical to, a previously mentioned entity.

In example~\ref{ex:bridging}, the antecedent is \emph{``February 1860''} and the anaphor is
\emph{``the following month''}, as the time in question cannot be interpreted without first interpreting the antecedent; therefore the correct answer is \textbf{yes}.

\ex. \label{ex:bridging}
\a.  \textit{Norton ordered all interested parties to assemble at Platt's Music Hall in San Francisco in \textbf{February 1860}. In an imperial decree \textbf{the following month}, Norton summoned the Army to depose the elected officials}
\b. \textbf{Detection}: \textit{yes}

\paragraph{Bridging type (\textsc{bridgtype})}
Given a bridging anaphor and its antecedent, the model must classify the type of semantic
relationship between them (e.g., \emph{entity-associative}, \emph{set-member}, \emph{part-whole}, etc.).
This task evaluates whether the model can recognize fine-grained associative links.
For this task, the input text highlights the antecedent using curly braces \{\} and the anaphor
using double dollar signs \$\$. We evaluate both \textbf{fine-grained} types and their corresponding \textbf{coarse} categories. (see Appendix~\ref{sec:label-appendix} for the full list of labels).
In example~\ref{ex:bridging} above, the cateogry is \textit{comparison-time}, since the identity of ``the following month'' can only be inferred by comparison to another time.

\paragraph{Bridging antecedent (\textsc{bridgante})} 
Given a marked bridging anaphor in a document, the model must identify its most likely
antecedent among the previous mentions in the discourse.  
In this task, the bridging anaphor is marked with curly braces \{\}, and the model is asked to
find the entity in the preceding text that serve as its antecedent. In the example above, only ``the following month'' would be highlighted, and the correct answer would be the string ``February 1960''.

\section{Evaluation}

To mitigate concerns about potential training data contamination, we note that several of our tasks are entirely new and unpublished or very recently published, including GUMBridge \citep{levine2025subjectivityannotationbridginganaphora} and the new datasets and labels introduced in DISRPT 2025, which postdate training cutoffs on our tested models. The modest model scores below also suggest LLMs have probably not seen or memorized this data.

\begin{table*}[t]
\centering
\scriptsize
\setlength{\tabcolsep}{3pt}
\renewcommand{\arraystretch}{1.1}

\scriptsize
\renewcommand{\arraystretch}{1.2}
\setlength{\tabcolsep}{4pt}

\begin{tabular}{lccccc}
  \toprule
  \multirow{2}{*}{\textbf{Model}} &
  \multirow{2}{*}{\textbf{English Avg}} &
  \multicolumn{3}{c}{\textbf{Multi-lingual Avg}} &
  \multirow{2}{*}{\textbf{All Tasks Avg}} \\
  \cmidrule(lr){3-5}
  & & \textbf{Eng} & \textbf{Non-Eng} & \textbf{Overall} & \\
  \midrule
  MT0 (zero) & 24.04 & 15.87 & 14.47 & 14.65 & 19.35 \\
  MT0 (few) & 20.39 & 12.07 & 11.85 & 11.75 & 16.07 \\
  Qwen (zero) & 35.80 & 20.92 & 17.10 & 17.05 & 26.42 \\
  Qwen (few) & 36.75 & 19.50 & 16.57 & 15.47 & 26.11 \\
  Aya (zero) & 37.66 & 16.16 & 13.87 & 14.13 & 25.90 \\
  Aya (few) & 38.17 & 17.96 & 14.58 & 15.83 & 27.00 \\
  GPT-4o mini (zero) & 42.54 & 36.63 & 34.13 & 34.62 & 38.58 \\
  GPT-4o mini (few) & 47.49 & \textbf{41.12} & 36.45 & 36.82 & 42.16 \\
  GPT-5 (zero) & 41.17 & 33.24 & 34.27 & 35.20 & 40.18 \\
  GPT-5 (few) & \textbf{51.79} & 38.46 & \textbf{36.83} & \textbf{36.90} & \textbf{44.34} \\
  \bottomrule
\end{tabular}
\label{tab:overall-performance}

\caption{Overall macro scores of each evaluated LLM.}
\label{tab:macro-average}
\end{table*}

\subsection{Human evaluation}

As shown in Figure~\ref{fig:llm-human-performance}, we conducted a human evaluation on all English subtasks by
asking two human annotators to solve 30 randomly selected examples for each task. The same subsets were used to evaluate model performance for a fair LLM--human comparison. While GPT-4o mini and GPT-5 obtain the best results among  the LLMs, human performance still surpasses all models across every subtask. 

At the same time, humans do not achieve perfect scores on any task, showing that even intuitive tasks such as counting mentions can depend on dataset-specific guidelines  (e.g.~mention definitions). Some tasks, such as inserting implicit connectives, also appear partly subjective, with humans scoring lowest there (though still above models).

\subsection{LLM performance}

Table~\ref{tab:macro-average} reports the overall macro F1 scores across all languages for each evaluated LLM.
Among the 11 subtasks in our benchmark, the three \textsc{bridg} tasks and the two \textsc{sal} tasks are English-only; the remaining six tasks are multilingual, and we report three aggregated scores for each model: All languages, English, and Non-English, where we expect English performance to be highest and non-English to be lowest.
For  \textsc{entcount} we report RMSE separately in the appendix, since it is not directly comparable to F1. This allows us to quantify how far model predictions are from the true number of mentions on average.

Overall, GPT-4o-mini and GPT-5 achieve the strongest results on most tasks, confirming their robustness across discourse-level evaluation, though as black box models, these results are not transparent or reproducible.
The exception is \textsc{bridgdet}, where Aya-Expanse-8B few-shot outperforms other models. Following recent work on benchmark correlation and meta-analysis, (e.g.~BenchBench, \citealt{zhang2024inherent}), we note that our model ranking largely agree with LMArena’s leaderboard \citep{chiang2024chatbot}, where GPT-5 is followed by GPT-4o-mini, Aya-Expanse-8B, and Qwen-1.5-7B, 
though specific tasks and performance gap sizes do vary from overall LMArena rankings.

\paragraph{Few-shot vs. zero-shot.}
In principle, few-shot prompting is expected to outperform zero-shot across tasks, since it can align models with the specific guidelines of the dataset, which is used to generate few-shot examples. However, our results show some notable deviations from this trend. In particular, Table~\ref{tab:pertask} shows that GPT-4o mini achieves substantially lower scores in the \textsc{bridgdet} few-shot setting than its zero-shot scores. We also observe smaller drops in a few other tasks. For instance, \textsc{bridgtype} and \textsc{entcount}, where few-shot underperforms zero-shot slightly. These exceptions aside, few-shot usually yields improvements or remains comparable to zero-shot.

\paragraph{English vs. non-English performance.}
For the multilingual tasks, one might expect English performance to be consistently higher than non-English due to model pretraining bias and data availability. However, our results reveal some interesting surprises: for \textsc{entcount} and \textsc{entmax}, non-English scores are higher for several models. 
For example, GPT-4o mini and GPT-5 both achieve around \textbf{60\%} F1 on Czech for \textsc{entmax}, and Russian and Spanish reach close to \textbf{50\%}, while English performance on the same task is around \textbf{30\%}. 
A similar trend can also be seen for \textsc{entcount}, where non-English often outperforms English. One possible reason may be that English data comes from a mix of spoken and written sources, while these language datasets are uniformly written. 
However the same is true for e.g.~Turkish, which is all written data but fares worse than English, and German, which is worse on \textsc{entmax} for all models (see the Appendix~\ref{sec:appendixB} for language specific results).

\subsection{Error Analysis}

\noindent
\begin{minipage}{0.48\textwidth} 
  \centering
  \includegraphics[width=\linewidth]{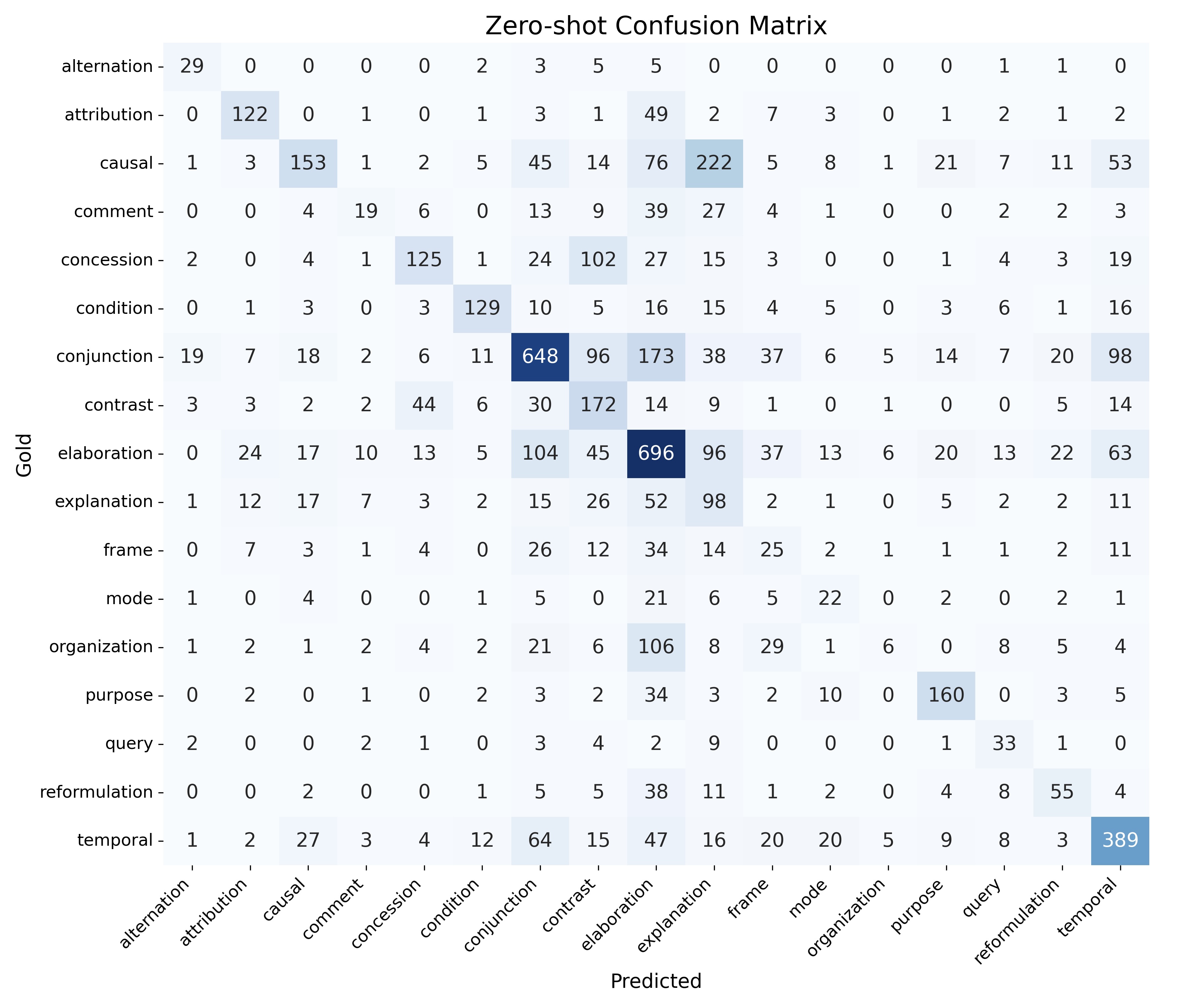}
  \captionof{figure}{Confusion matrix for the relation classification task in the zero-shot setting.}
  \label{fig:zero-confusion}
\end{minipage}

\paragraph{Relation classification confusions}
In the zero-shot setting (Figure~\ref{fig:zero-confusion}), GPT-5 shows high accuracy on \textsc{conjunction} and \textsc{elaboration}, likely because these frequent categories dominate the diagonal and are relatively easy to predict correctly.
At the same time, we observe substantial confusion between \textsc{causal} and \textsc{explanation} (the largest off-diagonal error, with 222 \textsc{causal} instances predicted as \textsc{explanation}), indicating the model is able to recognize relations which can be verbalized as `because', but does not distinguish well between objective causality (`Kim fell because Mary pushed her') and the explanatory type (`this hat will fit because I've had the same one before'). We also observe many confusions between \textsc{conjunction} and
\textsc{elaboration} or \textsc{temporal}, showing that while the model recognizes the additive nature of these relations, it does not distinguish well between subordinate elaborations and coordinate conjunctions (giving additional information rather than making a new point), and fails to capture whether we can implicitly know how two propositions are ordered in time.

In the few-shot setting (see Appendix~\ref{sec:appendixB} Figure~\ref{fig:confusion-matrices-all}), the model continues to show high accuracy on \textsc{conjunction} and \textsc{elaboration}; however, few-shot prompting does not outperform the zero-shot setting on these categories and in some cases even leads to slightly lower scores.
Some confusion patterns nevertheless remain: in particular, we still observe frequent confusion between \textsc{conjunction} and \textsc{organization}, as well as between \textsc{elaboration} and \textsc{organization}.

\paragraph{Bridging} The best performing task here is \textsc{bridgtype}, where few-shot prompting yields strong accuracy on \textsc{comparison} and \textsc{entity} (see Appendix \ref{sec:appendixB} Figure~\ref{fig:confusion-matrices-all}), while most of the remaining errors come from confusion between \textsc{comparison} and \textsc{set}. For smaller models, such as Qwen 7B, few-shot underperforms zero shot, perhaps because longer prompt create more distractions.

Human performance also shows that \textsc{comparison} and \textsc{entity} are the best-recognized categories, aligning with the model’s strengths, though model predictions of \textsc{set} for true \textsc{comparison} are proportionally much worse than for humans. Low performance on \textsc{bridgdet} and \textsc{bridgante}, at almost half the human scores even for GPT-5 few-shot, show that this remains a challenging task.

\section{Conclusion}

This paper presents DiscoTrack, a new multilingual benchmark with data from 12 diverse languages and a specific focus on tasks which target long-form, multi-sentential document understanding but are well-suited to a textual language generation format. Covering 11 subtasks in four areas (salience, entity tracking, discourse relations and bridging), our evaluation shows that even very large commercial models such as GPT-5 still struggle to reach human levels of performance. 

The results indicate that performance on English is often, but not always, better than on other languages, and we hypothesize that the greater diversity of English data in terms of genres and modality (spoken/written) may sometimes lead to higher scores outside of English, especially in easier data sources that rely on standard written language (for example Czech, Spanish and Russian data from CorefUD). We also find that while few-shot examples usually improve results, this is not always the case, especially for smaller models which can be distracted by longer prompts. Finally, we note that the prompts used for the evaluation were fairly simple, and we did not experiment with complex chain-of-thought prompts to improve LLM results on our tasks - we leave this and other more complex strategies to approaching human performance on our tasks to future work.

\section*{Limitations}

Due to license limitations, we were only able to include some of the datasets and languages made available in sources such as DISRPT and CorefUD -- results on other high quality datasets which could not be included may differ from our results. The selection of data is always subjective, and while we believe the breadth of our benchmark and the reliance on previous work make our scores meaningful, it is possible that systematic biases in some of our data may lead to misleading results. 

We are also mindful of the potential for LLM data contamination, but we note that many of the underlying resources are relatively new, and in some cases only just published in the past few months, meaning that the chance that models have seen the test data is overall low (as also evidenced by the modest performance of even very new models, such as GPT-5). 

Finally as noted in the paper, our naive prompting strategies included minimal guidelines to each task, the list of possible labels, and the same context and input data given to human evaluators. We did not experiment with more complex strategies, such as suggesting to models to serialize each sentence and count mentions of an entity only in it, then sum, or to construct rationales to support salience judgments. It is still possible that black box reasoning APIs as used by the GPT models tested here implement such strategies internally, but we speculate regardless that more prompt engineering could have resulted in higher performance for models. We view the main goal of this paper as introducing the new benchmark, and leave the goal of obtaining substantial improvements in LLM scores to future work.


\bibliographystyle{acl_natbib}
\bibliography{custom}

\appendix

\section{Prompts and Tag Sets}\label{sec:label-appendix}
This appendix provides the full set of prompts and tag definitions used in our benchmark tasks. 
Section~\ref{sec:label-appendix} first lists the exact prompts given to models for each task, 
followed by the tag sets used for classification. 
In total, we include 17 labels for core discourse relations, following the taxonomy of \citet{bunt2016iso}. 
For bridging anaphora, we adopt 11 fine-grained bridging types, as introduced by \citet{levine2025subjectivityannotationbridginganaphora}.
These materials are included to ensure transparency and reproducibility.

\subsection{Prompts}
\paragraph{SALTOP}
\begin{quote}
Which are the most salient entities in the text? \par
Here is the text:
\{text\} \\

\end{quote}

\paragraph{SALREL}
\begin{quote}
Which entity is more salient in the text: \{entity1\} or \{entity2\}? \par
Here is the text:
\{text\} \\ \\
\end{quote}

\paragraph{ENTMAX}
\begin{quote}
Which entity is mentioned the most times in the text? \par
Here is the text:
\{text\} \par

Ignore speaker labels (e.g., [Speaker: Kendra], [Speaker: Sabrina]). They are not entities and should not be considered as the answer. \end{quote}

\paragraph{ENTCOUNT}
\begin{quote}
How many times is \{entity\} mentioned in the text? \par
Here is the text:
\{text\} \par

Ignore speaker labels (e.g., [Speaker: Kendra], [Speaker: Sabrina]). They are not entities and should not be counted. \end{quote}

\paragraph{ENTCOREF}
\begin{quote}
In the following text, two mentions are marked with curly braces \{\}.
\par
Here is the text with two marked mentions:
\{text\} \\

Do the two marked mentions refer to the same entity?
Answer with "Yes" or "No".\end{quote}

\paragraph{RELCLF}
\begin{quote}
What is the discourse relation between Arg1\{arg1\} and Arg2\{arg2\}?
Arg1 and Arg2 are discourse units—short text spans that reflect the structure of the discourse.\par
You may also refer to their full sentences Text1\{text1\} and Text2\{text2\}
to better understand the context and make a more accurate decision.
Here are the options and definitions:
\begin{description}
  \item[alternation] One unit presents an option, and the other unit offers an alternative that is mutually exclusive with it.
  \item[attribution] One unit states some content, and the other gives information about its source.
  \item[causal] One unit presents a situation or event, and the other unit shows its cause or result.
  \item[comment] One unit expresses an opinion or evaluation about something stated in the other unit.
  \item[concession] One unit presents a situation that appears incompatible with the other unit, but the other unit remains valid.
  \item[condition] One unit presents a proposition and the other provides conditions that the other proposition depends on.
  \item[conjunction] One unit gives information, and the other unit continues with an additional point.
  \item[contrast] Two units present opposing or contrasting ideas.
  \item[elaboration] One unit adds extra details, explains, or further describes something introduced in the other unit.
  \item[explanation] One unit provides a reason, justification, or supporting information for what is stated in the other unit.
  \item[frame] One unit sets the scene or provides background, and the other unit presents the main information.
  \item[mode] One unit explains the way or method by which something in the other unit occurs.
  \item[organization] One unit helps structure the discourse, by introducing, preparing, or pointing out parts of the text.
  \item[purpose] One unit of the text describes an action or event, and the other unit explains the intended goal.
  \item[query] One unit presents a question or problem, and the other unit gives an answer or solution.
  \item[reformulation] One unit restates or rephrases the other unit in a different way.
  \item[temporal] One unit indicates that its event or situation happens before, after, or at the same time as the event in the other unit.
\end{description}
Please ignore the direction between Arg1 and Arg2 and focus on the overall relation between them.\par

\end{quote}

\paragraph{RELEXP}
\begin{quote}
An explicit discourse connective is a word or phrase which indicates the presence of a discourse relation between two segments, such as contrast, elaboration or temporal sequence. These connectives provide strong cues and allow us to classify discourse relations with high accuracy. Mark all explicit discourse connectives in the sentence using curly brackets. If there is no explicit discourse connective, answer with 'no explicit connective'.
Here are the options:
\textit{and, but, because,} \{... other options for this language\}

Here is the text:
\{text\}\par

\end{quote}

\paragraph{RELIMP}
\begin{quote}
What is the implicit discourse connective between \{arg1\} and \{arg2\}?
Implicit discourse connective refers to a logical connection between two discourse units (such as sentences or clauses) that is not explicitly marked by a discourse connective (e.g., “because”, “but”, “after”). Instead, the relation must be inferred by the reader or listener.
Here are the options: 
\textit{and, but, because,} \{... other options for this language\}

If there is no appropriate implicit connective, answer: no implicit connective.

\end{quote}

\paragraph{BRIDGDET}
\begin{quote}
Read the following text and answer with "yes" or "no": Does it contain a bridging anaphor?\par
Below is an explanation of what bridging and bridging anaphora are:
Bridging is an anaphoric phenomena which occurs when the reference of one entity is dependent on a previous associated, non-identical entity for interpretation. An instance of bridging consists of an antecedent-anaphor pair. The anaphor in this pair is referred to as the bridging anaphor. A bridging anaphor is a noun phrase that refers to something associated with, but not identical to, a previously mentioned entity. To understand a bridging anaphor, the reader must rely on their knowledge of the earlier entity. \par
Here is the text:
\{text\}
\end{quote}

\paragraph{BRIDGTYPE}

\begin{quote}
What is the bridging type between the anaphor and antecedent in the following text?

In the text, the antecedent is marked with curly braces \texttt{\{\}} and the anaphor is marked with double dollar signs \texttt{\$\$ \ \ \ \ \ \ \ \ \$\$}.

Below is an explanation of what bridging, bridging antecedents, and bridging anaphora are:

Bridging is an anaphoric phenomenon which occurs when the reference of one entity is dependent on a previous associated, non-identical entity for interpretation. An instance of bridging consists of an antecedent--anaphor pair.

The anaphor in this pair is referred to as the \emph{bridging anaphor}. A bridging anaphor is a noun phrase that refers to something associated with, but not identical to, a previously mentioned entity. To understand a bridging anaphor, the reader must rely on their knowledge of the earlier entity.

The \emph{bridging antecedent} provides the contextual basis or background knowledge that allows the reader or hearer to infer the meaning of the anaphor.

Here are the options and definitions:
\begin{description}
  \item[comparison-ellipsis] The anaphor is preceded by a comparative marker (other, another, same, more, ordinal modifiers, etc.) which implies a comparison to the antecedent.
  \item[comparison-sense] The type of the anaphor is omitted but inferable via comparison to the antecedent.
  \item[comparison-time] The anaphor refers to a specific time/timeframe which is understandable with reference to the antecedent.
  \item[entity-associative] The anaphor is an attribute or closely associated entity of the antecedent.
  \item[entity-meronymy] The anaphor is a subunit of the antecedent (part--whole).
  \item[entity-property] The anaphor is a physical or intangible property of the antecedent.
  \item[entity-resultative] The anaphor is logically inferable from the antecedent (e.g., result, transformation, cause).
  \item[set-member] The anaphor is an element of the antecedent set.
  \item[set-subset] The anaphor is a subset of the antecedent set.
  \item[set-span-interval] The anaphor is a sub-span of a spatial or temporal interval defined by the antecedent.
  \item[other] Valid bridging pair but does not fall under the above types.\par

\end{description}

\noindent\textbf{Note:} Bridging types are symmetric with respect to anaphor and antecedent. For example, both ``a member of the team'' and ``the team'' referring to each other are labeled as \texttt{set-member}. Similarly, both ``the capital'' and ``the country'' can be \texttt{entity-associative}, regardless of order.\par

Here is the text with  \texttt{\{\}} and  \texttt{\$\$ \$\$} :\{text\}

\end{quote}  

\paragraph{BRIDGANTE}
\begin{quote}
In the following sentence, the bridging anaphora is marked with curly brackets. What is its antecedent(s)?\par
Below is an explanation of what bridging, bridging antecedents, and bridging anaphora are:
Bridging is an anaphoric phenomena which occurs when the reference of one entity is dependent on a previous associated, non-identical entity for interpretation. An instance of bridging consists of an antecedent-anaphor pair.
The anaphor in this pair is referred to as the bridging anaphor. A bridging anaphor is a noun phrase that refers to something associated with, but not identical to, a previously mentioned entity. To understand a bridging anaphor, the reader must rely on their knowledge of the earlier entity.
The bridging antecedent provides the contextual basis or background knowledge that allows the reader or hearer to infer the meaning of the anaphor. \par
Here is the text:
\{text\}
\end{quote}

\subsection{Tag Sets}

\paragraph{Bridging Type}
\begin{itemize}[leftmargin=*,itemsep=0pt,topsep=0pt]
  \item comparison-ellipsis
  \item comparison-sense
  \item comparison-time
  \item entity-associative
  \item entity-meronymy
  \item entity-property
  \item entity-resultative
  \item set-member
  \item set-subset
  \item set-span-interval
  \item other
\end{itemize}

\paragraph{Discourse Relation}
\begin{itemize}[leftmargin=*,itemsep=0pt,topsep=0pt]
  \item alternation
  \item attribution
  \item causal
  \item comment
  \item concession
  \item condition
  \item conjunction
  \item contrast
  \item elaboration
  \item explanation
  \item frame
  \item mode
  \item organization
  \item purpose
  \item query
  \item reformulation
  \item temporal
\end{itemize}

\noindent\emph{Definitions of these tags are provided in the task prompt above.}

\section{Additional Figures and Tables} \label{sec:appendixB}

\begin{figure*}[t]
\centering

\subfloat[Discourse relation (few-shot)]{%
    \includegraphics[width=0.45\textwidth]{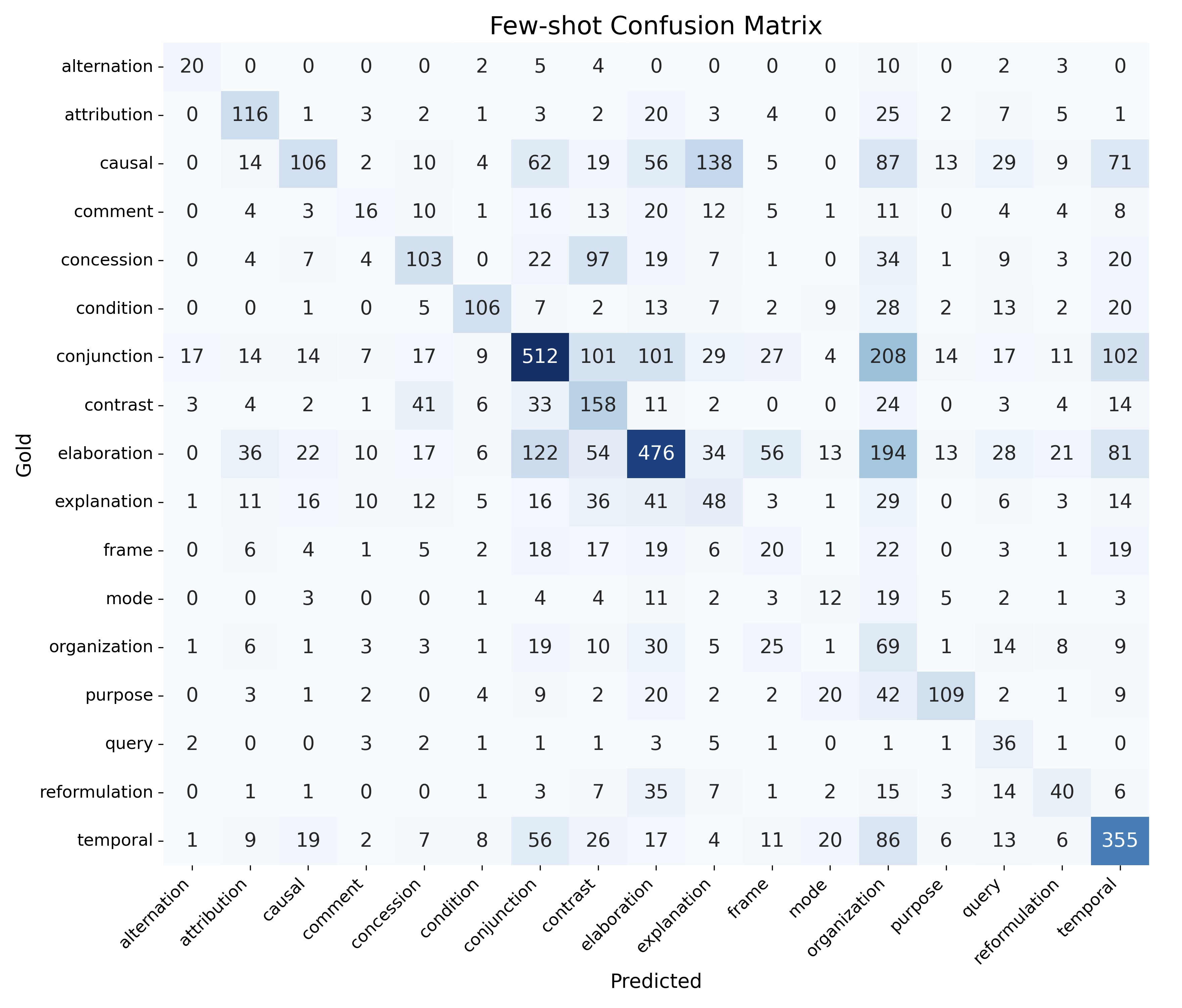}
}\hfill
\subfloat[Bridging type (zero-shot)]{%
    \includegraphics[width=0.45\textwidth]{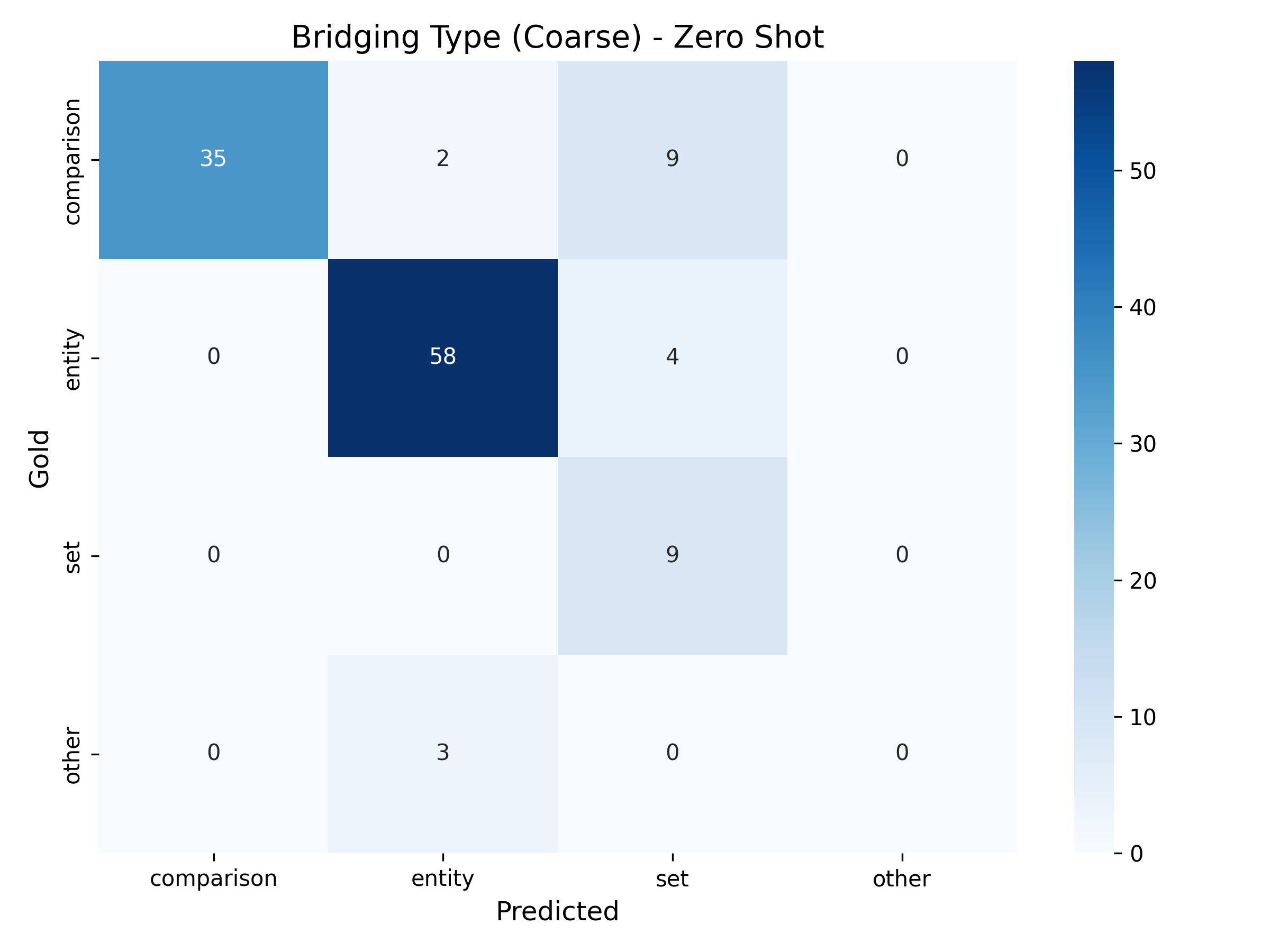}
}

\vspace{0.5em}
\subfloat[Bridging type (few-shot)]{%
    \includegraphics[width=0.45\textwidth]{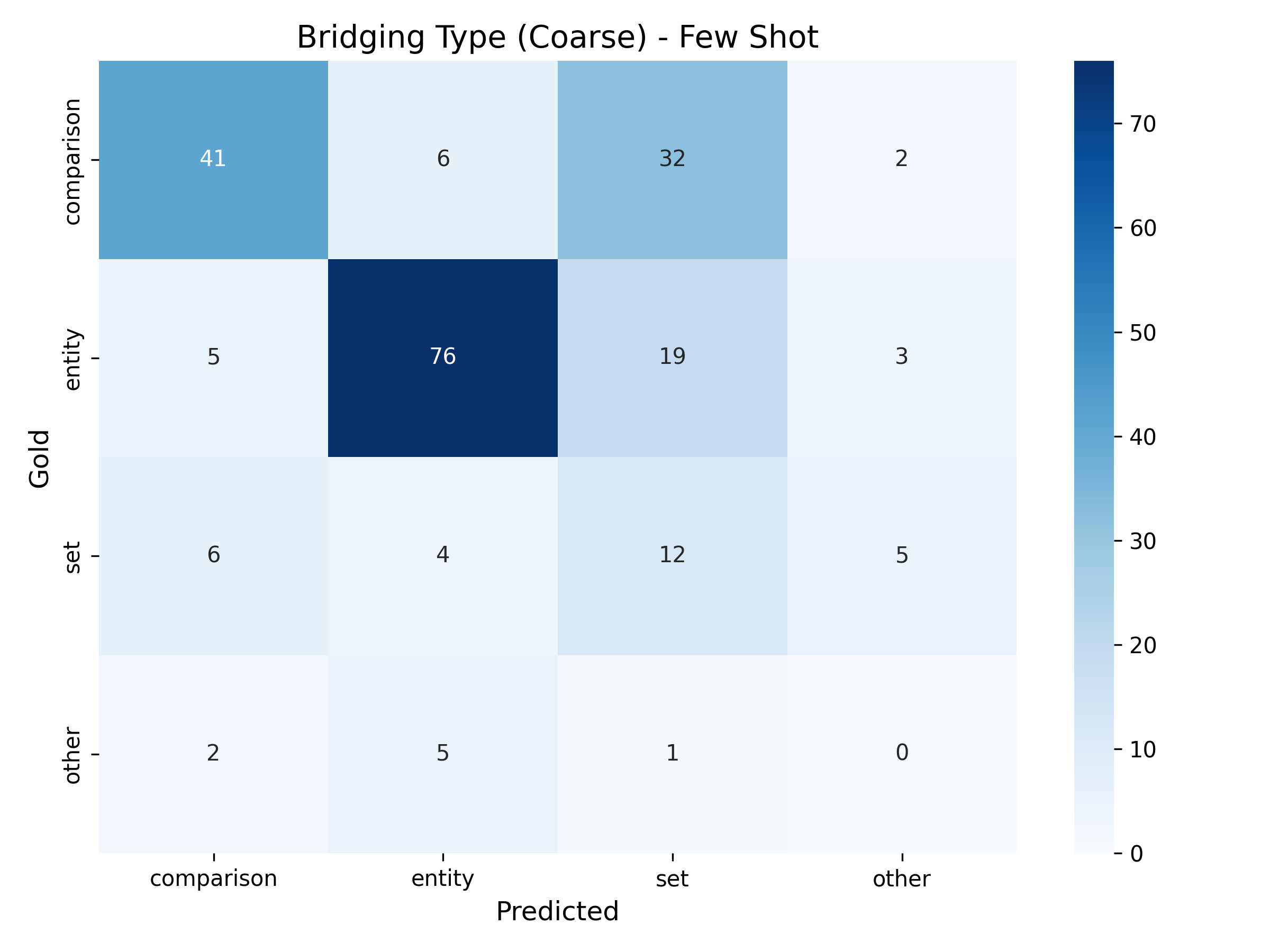}
}\hfill
\subfloat[Human performance]{%
    \includegraphics[width=0.45\textwidth]{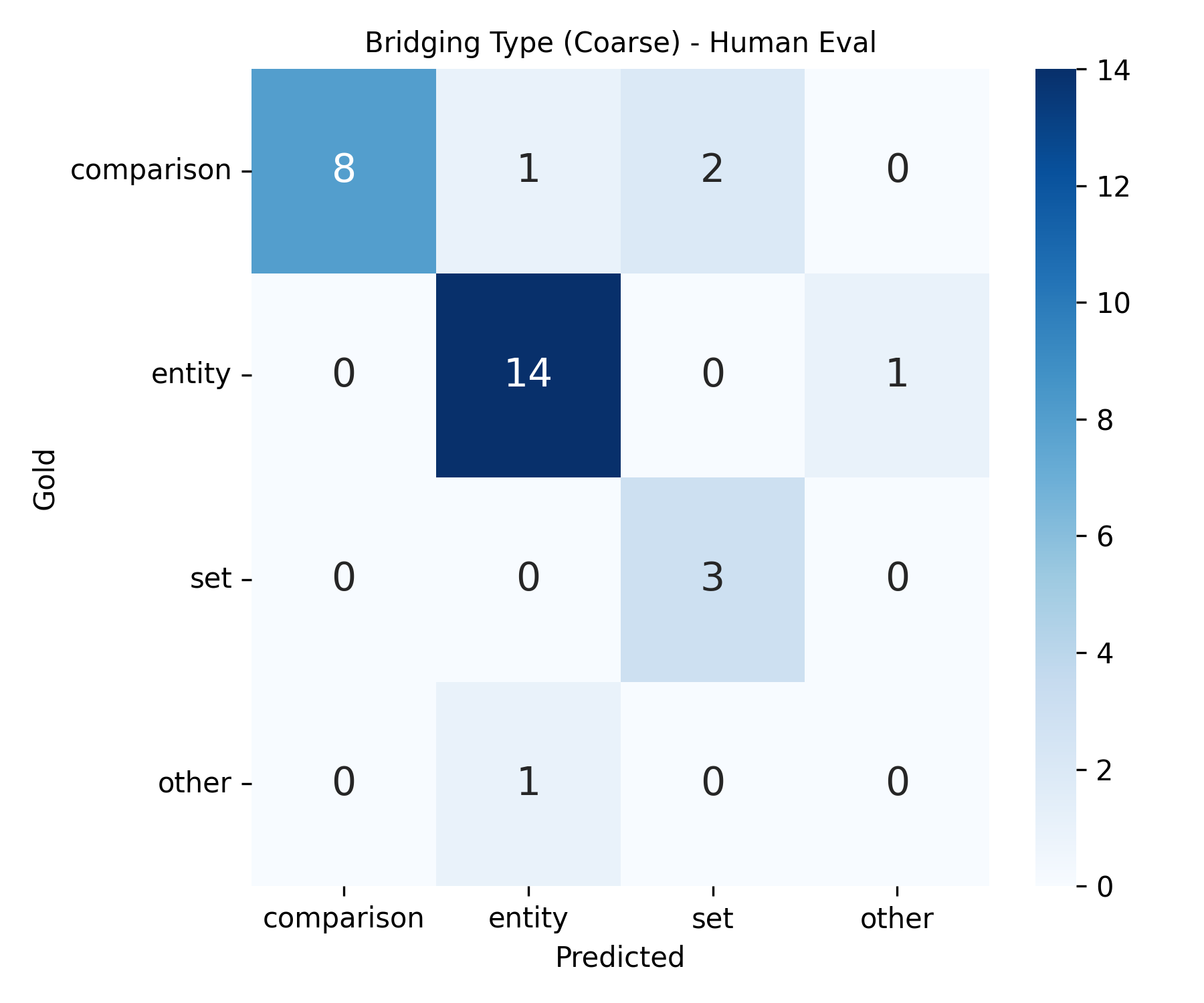}
}

\caption{Confusion matrices for (a) discourse relation classification in the few-shot setting, (b) bridging type classification in the zero-shot setting, (c) bridging type classification in the few-shot setting, and (d) human performance on bridging type classification.}
\label{fig:confusion-matrices-all}
\end{figure*}

As shown in Figure~\ref{fig:confusion-matrices-all}, we provide four additional confusion matrices: 
discourse relation classification in the few-shot setting, 
bridging type classification in the zero-shot and few-shot settings, 
and human performance on bridging type classification.
We also include supplementary tables 
(Table~\ref{tab:eng_only_results} through Table~\ref{tab:divide})
reporting per-task and per-language scores for all evaluated models, 
including detailed accuracy and F1 metrics. 
For the Entity Count task, we additionally report RMSE separately.

\begin{table*}[ht]
\centering
\scriptsize
\renewcommand{\arraystretch}{1.2}
\setlength{\tabcolsep}{3pt}
\begin{adjustbox}{max width=\textwidth}
\begin{tabular}{lcccccc}
\toprule
\textbf{Model} & \textbf{BRIDGTYPE} & \textbf{BRIDGDET} & \textbf{BRIDGANTE} & \textbf{SALREL} & \textbf{SALTOP} \\
\midrule
MT0 (zero) & 6.85 / 12.33 & 45.49 & 2.05 & 54.20 & 23.30 \\
MT0 (few)  & 1.37 / 47.03 & 50.21 & 5.13 & 13.80 & 4.80 \\
Qwen (zero) & 30.14 / 47.95 & 48.28 & 0.51 & 52.80 & 35.10 \\
Qwen (few)  & 15.98 / 49.40 & 52.15 & 2.56 & 59.20 & 41.22 \\
Aya (zero) & 29.68 / 47.03 & 50.21 & 8.21 & 52.20 & 38.65 \\
Aya (few)  & 29.69 / 47.03 & \textbf{62.23} & 1.03 & 49.60 & 39.45 \\
GPT-4o-mini (zero) & \textbf{32.42} / 54.34 & 57.30 & 18.97 & 52.60 & 39.61 \\
GPT-4o-mini (few)  & 30.59 / 52.97 & 49.14 & \textbf{42.05} & 67.00 & 43.16 \\
GPT-5 (zero) & 21.92 / 46.56 & 53.65 & 38.46 & 54.2 & 56.2 \\
GPT-5 (few)  & 31.51 / \textbf{58.9} & 54.29 & 38.97 & \textbf{67.8} & \textbf{59.26} \\
\bottomrule
\end{tabular}
\end{adjustbox}
\caption{Results for five English-only tasks.}
\label{tab:eng_only_results}
\end{table*}

\begin{table*}[ht]
\centering
\scriptsize
\renewcommand{\arraystretch}{1.2}
\setlength{\tabcolsep}{2pt}
\begin{adjustbox}{max width=\textwidth}
\begin{tabular}{l
cc cc cc cc cc cc cc cc cc cc cc cc cc}
\toprule
\multirow{2}{*}{\textbf{Model}} 
& \multicolumn{2}{c}{English} 
& \multicolumn{2}{c}{German} 
& \multicolumn{2}{c}{Dutch} 
& \multicolumn{2}{c}{Turkish} 
& \multicolumn{2}{c}{Czech} 
& \multicolumn{2}{c}{French} 
& \multicolumn{2}{c}{Russian} 
& \multicolumn{2}{c}{Spanish} \\
\cmidrule(lr){2-3}\cmidrule(lr){4-5}\cmidrule(lr){6-7}\cmidrule(lr){8-9}
\cmidrule(lr){10-11}\cmidrule(lr){12-13}\cmidrule(lr){14-15}\cmidrule(lr){16-17}
& Zero & Few & Zero & Few & Zero & Few & Zero & Few & Zero & Few & Zero & Few & Zero & Few & Zero & Few \\
\midrule
MT0       & 28.21 & 2.99  & 9.50 & 3.17   & 18.60 & 10.20 & 13.64 & 4.55  & 23.00 & 0.40  & 16.78 & 0.00  & 34.64 & 0.00  & 35.40 & 0.20  \\
Qwen      & 22.22 & 23.50 & 3.62 & 6.33   & 10.00 & 16.20 & 4.55  & 13.64 & 12.80 & 8.60  & 11.89 & 0.00  & 15.64 & 0.00  & 27.20 & 26.40 \\
Aya       & 25.21 & 29.91 & 1.81 & 1.81   & 1.40  & 1.00  & 0.00  & 0.00  & 15.20 & 22.00 & 2.10  & 5.59  & 34.08 & 35.20 & 0.80  & 1.20  \\
GPT-4o-mini    & 31.62 & \textbf{35.04} & 23.52 & 19.46 & \textbf{29.20} & \textbf{31.80} & 18.20 & \textbf{31.82} & 63.40 & \textbf{66.20} & \textbf{34.27} & \textbf{32.17} & 56.42 & \textbf{59.78} & 53.20 & \textbf{55.60} \\
GPT-5    & \textbf{37.18} & 32.91 & \textbf{26.24} & \textbf{25.79} & 28.2 & \textbf{31.80} & \textbf{22.73} & 27.27 & \textbf{70.4} & 45.8 & 32.17 & 27.97 & \textbf{58.66} & 50.84 & \textbf{57.2} & 45.4 \\
\bottomrule
\end{tabular}
\end{adjustbox}
\caption{Results for \textbf{ENTMAX} across 8 languages.}
\end{table*}

\begin{table*}[ht]
\centering
\scriptsize
\renewcommand{\arraystretch}{1.2}
\setlength{\tabcolsep}{2pt}
\begin{adjustbox}{max width=\textwidth}
\begin{tabular}{l
cc cc cc cc cc cc cc cc cc cc cc cc cc}
\toprule
\multirow{2}{*}{\textbf{Model}} 
& \multicolumn{2}{c}{English} 
& \multicolumn{2}{c}{German} 
& \multicolumn{2}{c}{Polish} 
& \multicolumn{2}{c}{Turkish} 
& \multicolumn{2}{c}{Czech} 
& \multicolumn{2}{c}{French} 
& \multicolumn{2}{c}{Russian} 
& \multicolumn{2}{c}{Spanish} \\
\cmidrule(lr){2-3}\cmidrule(lr){4-5}\cmidrule(lr){6-7}\cmidrule(lr){8-9}
\cmidrule(lr){10-11}\cmidrule(lr){12-13}\cmidrule(lr){14-15}\cmidrule(lr){16-17}
& Zero & Few & Zero & Few & Zero & Few & Zero & Few & Zero & Few & Zero & Few & Zero & Few & Zero & Few \\
\midrule
MT0       & 0.74  & 0.74  & 1.73  & 18.50 & 3.80  & 1.40  & \textbf{5.60}  & 0.00  & 4.00  & 0.50  & 0.00  & 0.00  & \textbf{5.90}  & 5.36  & 1.06  & \textbf{6.36}  \\
Qwen      & 5.19  & 8.40  & 17.92 & 19.08 & 5.40  & 4.20  & 4.52  & 3.72  & 2.00  & 0.00  & 6.05  & 0.00  & 9.12  & \textbf{5.36}  & 2.97  & 5.08  \\
Aya       & 3.95  & 5.19  & 5.78  & 8.67  & 0.20  & 0.00  & 0.80  & \textbf{7.71}  & \textbf{5.20}  & \textbf{5.00}  & 2.91  & 0.00  & 1.61  & 0.00  & 1.69  & 2.12  \\
GPT-4o-mini    & \textbf{10.37} & \textbf{11.60} & \textbf{23.12} & 20.23 & \textbf{10.40} & \textbf{11.20} & 5.05  & 7.18  & 3.00  & 2.40  & \textbf{7.75}  & 2.18  & 5.09  & 5.00  & \textbf{5.30}  & 4.66  \\
GPT-5    & \textbf{10.37} & 11.36 & 20.81 & \textbf{20.81} & 10.2 & 9.6 & 4.52 & 5.32  & 2.6  & 2  & 7.02  & \textbf{7.26}  & 4.83  & 4.29  & 4.45  & 4.03  \\
\bottomrule
\end{tabular}
\end{adjustbox}
\caption{Results for \textbf{ENTCOUNT} across 8 languages.}
\end{table*}

\begin{table*}[ht]
\centering
\scriptsize
\renewcommand{\arraystretch}{1.2}
\setlength{\tabcolsep}{2pt}
\begin{adjustbox}{max width=\textwidth}
\begin{tabular}{l
cc cc cc cc cc cc cc cc cc cc cc cc cc}
\toprule
\multirow{2}{*}{\textbf{Model}} 
& \multicolumn{2}{c}{English} 
& \multicolumn{2}{c}{German} 
& \multicolumn{2}{c}{Polish} 
& \multicolumn{2}{c}{Turkish} 
& \multicolumn{2}{c}{Czech} 
& \multicolumn{2}{c}{French} 
& \multicolumn{2}{c}{Russian} 
& \multicolumn{2}{c}{Spanish} \\
\cmidrule(lr){2-3}\cmidrule(lr){4-5}\cmidrule(lr){6-7}\cmidrule(lr){8-9}
\cmidrule(lr){10-11}\cmidrule(lr){12-13}\cmidrule(lr){14-15}\cmidrule(lr){16-17}
& Zero & Few & Zero & Few & Zero & Few & Zero & Few & Zero & Few & Zero & Few & Zero & Few & Zero & Few \\
\midrule
MT0       & 25.25 & 9.20 & 45.24 & \textbf{2.75} & 83.29 & 10.21 & 101.26 & 9.82 & >1000 & 10.26 & 10.53 & 10.31 & 113.03 & \textbf{5.47} & 95.58 & \textbf{5.58} \\
Qwen      & >1000 & 10.60 & 23.39 & 4.00 & 60.41 & 17.10 & 64.06 & 8.72 & >1000 & 25.17 & 9.03 & 10.32 & 9.14 & 8.56 & 10.34 & 9.79 \\
Aya       & 139.26 & 6.61 & 721.90 & 3.18 & >1000 & 11.91 & 814.70 & \textbf{7.40} & >1000 & 89.3 & >1000 & >1000 & >1000 & >1000 & 738.59 & 10.52 \\
GPT-4o-mini    & \textbf{7.46} & \textbf{8.00} & \textbf{3.46} & 3.82 & \textbf{9.86} & \textbf{9.41} & 8.23 & 7.62 & \textbf{9.26} & \textbf{9.88} & 9.34 & 9.43 & \textbf{6.91} & 7.45 & \textbf{8.15} & 9.20 \\
GPT-5    & 8.25 & 8.07 & 3.68 & 3.77 & 9.97 & \textbf{9.41} & \textbf{8.09} & 8.21 & 10.43 & 10.34 & \textbf{9.29} & \textbf{8.99} & 7.49 & 7.74 & 9.45 & 9.76 \\
\bottomrule
\end{tabular}
\end{adjustbox}
\caption{RMSE Results for \textbf{ENTCOUNT} task across 8 languages.}
\end{table*}  

\begin{table*}[ht]
\centering
\scriptsize
\renewcommand{\arraystretch}{1.2}
\setlength{\tabcolsep}{2pt}
\begin{adjustbox}{max width=\textwidth}
\begin{tabular}{l
cc cc cc cc cc cc cc cc cc cc cc cc cc}
\toprule
\multirow{2}{*}{\textbf{Model}} 
& \multicolumn{2}{c}{English} 
& \multicolumn{2}{c}{German} 
& \multicolumn{2}{c}{Polish} 
& \multicolumn{2}{c}{Turkish} 
& \multicolumn{2}{c}{Czech} 
& \multicolumn{2}{c}{French} 
& \multicolumn{2}{c}{Russian} 
& \multicolumn{2}{c}{Spanish} \\
\cmidrule(lr){2-3}\cmidrule(lr){4-5}\cmidrule(lr){6-7}\cmidrule(lr){8-9}
\cmidrule(lr){10-11}\cmidrule(lr){12-13}\cmidrule(lr){14-15}\cmidrule(lr){16-17}
& Zero & Few & Zero & Few & Zero & Few & Zero & Few & Zero & Few & Zero & Few & Zero & Few & Zero & Few \\
\midrule
MT0       & 38.40 & 49.00 & 63.46 & 71.75 & 38.20 & 44.60 & 31.80 & 30.80 & 39.40 & 49.60 & 26.60 &  6.00 & 34.20 & 50.00 & 33.40 & 41.20 \\
Qwen      & 55.80 & 41.40 & 34.62 & 27.88 & 51.40 & 44.80 & 59.80 & 22.60 & 59.80 & 22.60 & 88.80 & 87.40 & 50.80 & 50.80 & 58.40 & 63.00 \\
Aya       & 54.20 & 50.00 & 25.96 & 21.15 & 51.00 & 50.00 & 53.60 & 54.00 & 52.20 & 50.00 & \textbf{93.80} & \textbf{93.80} & 53.60 & 50.00 & \textbf{58.60} & 43.00 \\
GPT-4o-mini    & \textbf{76.00} & \textbf{81.60} & 85.58 & 76.92 & \textbf{81.00} & \textbf{78.60} & \textbf{67.60} & 61.40 & \textbf{75.20} & \textbf{80.00} & 41.60 & 73.20 & \textbf{76.80} & 75.40 & 53.80 & \textbf{56.60} \\
GPT-5    & 72 & 75.4 & \textbf{87.5} & \textbf{80.77 }& 73.6 & 80.8 & 62.2 & \textbf{64.6} & 60.4 & 79.2 & 19.6 & 75.8 & 63.2 & \textbf{77} & 47.2 & 52.2 \\
\bottomrule
\end{tabular}
\end{adjustbox}
\caption{Results for \textbf{ENTCOREF } across 8 languages.}
\end{table*}

\begin{table*}[ht]
\centering
\scriptsize
\renewcommand{\arraystretch}{1.2}
\setlength{\tabcolsep}{2pt}
\begin{adjustbox}{max width=\textwidth}
\begin{tabular}{l
cc cc cc cc cc cc cc cc cc cc cc cc cc cc cc cc cc}
\toprule
\multirow{2}{*}{\textbf{Model}} 
& \multicolumn{2}{c}{English} 
& \multicolumn{2}{c}{German} 
& \multicolumn{2}{c}{Polish} 
& \multicolumn{2}{c}{Portuguese} 
& \multicolumn{2}{c}{Turkish} 
& \multicolumn{2}{c}{Czech} 
& \multicolumn{2}{c}{Chinese} 
& \multicolumn{2}{c}{Italian} 
& \multicolumn{2}{c}{Thai} \\
\cmidrule(lr){2-3}\cmidrule(lr){4-5}\cmidrule(lr){6-7}\cmidrule(lr){8-9}
\cmidrule(lr){10-11}\cmidrule(lr){12-13}\cmidrule(lr){14-15}\cmidrule(lr){16-17}\cmidrule(lr){18-19}
& Zero & Few & Zero & Few & Zero & Few & Zero & Few & Zero & Few & Zero & Few & Zero & Few & Zero & Few & Zero & Few \\
\midrule
MT0       & 0.00  & 0.00  & 0.00  & 0.00  & 0.00  & 0.00  & 0.00  & 0.00  & 0.00  & 0.00  & 0.00  & 1.05  & 0.00  & 0.00  & 0.00  & 0.00  & 0.00  & 0.00  \\
Qwen      &16.93  &18.32  &10.91  &36.78  & 3.15  & 6.44  & 2.79  &13.74  & 3.46  & 2.22  & 5.08  &29.14  & 8.51  &17.45  & 4.86  &18.14  & 1.05  & 2.33  \\
Aya       & 0.00  & 0.00  & 0.00  &45.10  & 0.00  & 0.00  & 0.00  & 0.00  & 0.00  & 0.00  & 0.00  & 0.00  &12.28  &32.31  &15.32  &51.06  & 0.00  & 0.00  \\
GPT-4o-mini    &\textbf{47.18}  &\textbf{52.70}  &57.97  &\textbf{70.90}  &\textbf{40.21}  &\textbf{42.15}  &\textbf{31.69}  &\textbf{36.72}  &43.94  &40.36  &\textbf{48.10}  &52.68  &32.31  &\textbf{44.35}  &51.06  &51.61  &\textbf{11.49}  &\textbf{11.49}  \\
GPT-5    &46.98  &47.18  &\textbf{60.7}  &65.19  &33.07  &37.33  &26.6  &32.07  &\textbf{44.36}  &\textbf{45.72}  &46.97  &\textbf{48.21}  &\textbf{33.79}  &39.84  &\textbf{56.26}  &\textbf{61.22}  &7.56  &9.79  \\
\bottomrule
\end{tabular}
\end{adjustbox}
\caption{Results for \textbf{RELEXP } across 9 languages.}
\end{table*}

\begin{table*}[ht]
\centering
\scriptsize
\renewcommand{\arraystretch}{1.2}
\setlength{\tabcolsep}{2pt}
\begin{adjustbox}{max width=\textwidth}
\begin{tabular}{l
cc cc cc cc cc cc cc cc cc cc cc cc}
\toprule
\multirow{2}{*}{\textbf{Model}} 
& \multicolumn{2}{c}{English} 
& \multicolumn{2}{c}{German} 
& \multicolumn{2}{c}{Polish} 
& \multicolumn{2}{c}{Portuguese} 
& \multicolumn{2}{c}{Turkish} 
& \multicolumn{2}{c}{Chinese} 
& \multicolumn{2}{c}{Russian}\\
\cmidrule(lr){2-3}\cmidrule(lr){4-5}\cmidrule(lr){6-7}\cmidrule(lr){8-9}\cmidrule(lr){10-11}\cmidrule(lr){12-13}\cmidrule(lr){14-15}
& Zero & Few & Zero & Few & Zero & Few & Zero & Few & Zero & Few & Zero & Few &Zero & Few\\
\midrule
MT0       &18.6 & 12.6 &\textbf{29.24} &\textbf{28.90} &30.37 &\textbf{38.34} &17.72 &14.56 &28.07 & 1.75 & 7.20 & 1.60 & 8.11 & 23.31\\
Qwen      &14.8 &15 &18.27 & 6.98 &\textbf{31.90} & 5.52 & 8.23 & 7.59 & 4.09 & 4.39 &18.20 & 4.60 & 3.72 & 3.72 \\
Aya       &8.2 &0.4 &17.67 & 2.66 & 8.28 & 5.21 &13.92 & 0.32 &14.91 &15.20 & 5.80 & 0.20 & \textbf{23.31} & 4.05\\
GPT-4o-mini    &21 &24.4 &14.62 &19.60 &13.80 &14.72 &\textbf{18.67} &\textbf{19.62} &15.20 &\textbf{26.90} & 5.41 &17.57 & 9.12 & 13.51\\
GPT-5    &\textbf{21.2} &\textbf{24.4} &13.95 &24.92 &8.9 &21.17 &1.84 &12.58 &8.48 &20.18 &12.6 &\textbf{39.8} & 15.2 & \textbf{20.27}\\
\bottomrule
\end{tabular}
\end{adjustbox}
\caption{Results for \textbf{RELIMP} across 7 languages.}
\end{table*}

\begin{table*}[ht]
\centering
\scriptsize
\renewcommand{\arraystretch}{1.2}
\setlength{\tabcolsep}{2pt}
\begin{adjustbox}{max width=\textwidth}
\begin{tabular}{l
cc cc cc cc cc cc cc cc cc cc cc cc cc cc cc cc cc cc cc cc cc cc cc}
\toprule
\multirow{2}{*}{\textbf{Model}} 
& \multicolumn{2}{c}{English} 
& \multicolumn{2}{c}{German} 
& \multicolumn{2}{c}{Polish} 
& \multicolumn{2}{c}{Portuguese} 
& \multicolumn{2}{c}{Turkish} 
& \multicolumn{2}{c}{Czech} 
& \multicolumn{2}{c}{Chinese} 
& \multicolumn{2}{c}{Italian} 
& \multicolumn{2}{c}{Thai} 
& \multicolumn{2}{c}{French} 
& \multicolumn{2}{c}{Russian} 
& \multicolumn{2}{c}{Spanish} \\
\cmidrule(lr){2-3}\cmidrule(lr){4-5}\cmidrule(lr){6-7}\cmidrule(lr){8-9}
\cmidrule(lr){10-11}\cmidrule(lr){12-13}\cmidrule(lr){14-15}\cmidrule(lr){16-17}
\cmidrule(lr){18-19}\cmidrule(lr){20-21}\cmidrule(lr){22-23}\cmidrule(lr){24-25}
& Zero & Few & Zero & Few & Zero & Few & Zero & Few & Zero & Few & Zero & Few & Zero & Few & Zero & Few & Zero & Few & Zero & Few & Zero & Few & Zero & Few \\
\midrule
MT0       &  2.60 &  0.40 &  3.00 &  5.40 &  3.00 & 11.16 &  6.80 &  0.60 &  2.20 &  0.80 &  5.40 &  7.00 &  0.40 &  0.00 &  1.00 & 11.20 &  1.00 & 11.20 &  3.20 &  0.80 &  5.80 &  0.60 & 17.40 &  0.00 \\
Qwen      & 10.60 & 10.40 & 11.20 & 11.40 & 11.20 & 11.40 &  7.00 &  9.20 &  6.00 &  3.60 &  4.40 &  1.80 &  1.60 &  1.20 &  6.00 &  1.40 &  6.00 &  1.40 &  5.80 &  2.60 &  4.00 &  4.20 &  1.60 &  1.20 \\
Aya       &  5.40 & 21.20 &  0.60 &  9.20 &  3.40 & 22.00 &  4.20 & 12.60 &  3.60 & 16.20 &  4.40 & 12.80 &  2.20 &  0.00 &  0.60 & 22.00 &  0.60 & 22.00 &  4.60 & 26.40 &  1.40 &  7.00 &  2.20 &  0.00 \\
GPT-4o-mini    & 33.40 & \textbf{41.10} & 35.60 & \textbf{36.20} & 48.40 & \textbf{48.20} & 54.00 & \textbf{54.20} & 41.40 & 39.20 &  9.80 & 10.40 &  6.40 &  7.80 & 10.40 &  9.00 & 10.40 &  9.00 & 10.60 & 11.20 & 40.20 & 38.80 &  6.40 &  7.80 \\
GPT-5    & \textbf{45.8} & 40.2 & \textbf{43.00} & 33.6 & \textbf{50.8} & 32.6 & \textbf{62} & 52.2 & \textbf{43.2} & \textbf{34.6} & \textbf{42.8} & \textbf{36.8} &  \textbf{41.6} & \textbf{34.6} & \textbf{53} &  \textbf{49.4} & \textbf{42.8} & \textbf{29.8} & \textbf{42} & \textbf{38.8} & \textbf{65.4} & \textbf{45.8} & \textbf{45.8} &  \textbf{32} \\
\bottomrule
\end{tabular}
\end{adjustbox}
\caption{Results for \textbf{RELCLF} across 12 languages.}
\label{tab:divide}
\end{table*}

\end{document}